\documentclass[conference]{IEEEtran}
\IEEEoverridecommandlockouts
\usepackage{cite}
\usepackage{hyperref}
\usepackage{multirow}
\hypersetup{
    colorlinks=true,
    linkcolor=black,
    citecolor=black,
    urlcolor=black,
}
\usepackage{amsmath,amssymb,amsfonts}
\usepackage{algorithmic}
\usepackage{graphicx}
\usepackage{textcomp}
\usepackage{xcolor}
\def\BibTeX{{\rm B\kern-.05em{\sc i\kern-.025em b}\kern-.08em
    T\kern-.1667em\lower.7ex\hbox{E}\kern-.125emX}}
\begin{document}

\title{ERUPD - English to Roman Urdu Parallel Dataset}
\author{
    \IEEEauthorblockN{Mohammed Furqan*\thanks{
        \hrule\vspace{1mm} 
        \textit{*Department of Computer Science and Engineering, Muffakham Jah College of Engineering and Technology}
    }}
    \IEEEauthorblockA{
        Hyderabad, India \\
        furqan02md@gmail.com
    }
    \and
    \IEEEauthorblockN{Raahid Bin Khaja*}
    \IEEEauthorblockA{
        Hyderabad, India \\
        raahiduae@gmail.com
    }
    \and
    \IEEEauthorblockN{Rayyan Habeeb*}
    \IEEEauthorblockA{
        Hyderabad, India \\
        rayyanhabeeb0@gmail.com
    }
}

\maketitle

\begin{abstract}
Bridging linguistic gaps fosters global growth and cultural exchange. This study addresses the challenges of Roman Urdu---a Latin-script adaptation of Urdu widely used in digital communication---by creating a novel parallel dataset comprising 75,146 sentence pairs. Roman Urdu's lack of standardization, phonetic variability, and code-switching with English complicates language processing. We tackled this by employing a hybrid approach that combines synthetic data generated via advanced prompt engineering with real-world conversational data from personal messaging groups. We further refined the dataset through a human evaluation phase, addressing linguistic inconsistencies and ensuring accuracy in code-switching, phonetic representations, and synonym variability. The resulting dataset captures Roman Urdu's diverse linguistic features and serves as a critical resource for machine translation, sentiment analysis, and multilingual education.
\end{abstract}


\section{Introduction}
Language proficiency drives social cohesion, economic development, and cultural exchange in today's interconnected society. Language barriers prevent speakers of different languages from communicating effectively, particularly for languages with different scripts and orthographies. With more than 170 million speakers worldwide\cite{shahroz2020rutut,begum2022national}, Urdu is traditionally written in a modified Persian-Arabic script. However, in digital contexts, it is commonly represented in Latin script, or "Roman Urdu."Roman Urdu has grown in popularity in online and mobile messaging platforms, where Urdu speakers utilize Latin characters to communicate more effectively. Nowadays, Urdu-speaking communities, especially the younger generation, use this digital vernacular extensively on social media, messaging apps, and other unofficial online forums. Despite its increasing popularity, language processing tools still struggle with Roman Urdu, and considerable resource gaps prevent researchers from effectively integrating it into natural language processing (NLP) systems.

Roman Urdu's extensive phonetic variability, high reliance on contextual meaning, and lack of standardized spelling conventions make it difficult to translate and interpret, even amongst the Urdu-speaking populace.  Speakers frequently create unique spelling based on individual phonetic approximations, which differ greatly from formally written Urdu.Roman Urdu's informal nature makes it prone to slang, colloquial expressions, and code-switching with English. The unavailability of Roman Urdu to English translation systems creates significant challenges, particularly in identifying hate speech and toxicity within online communities. These challenges multiply due to the scarcity of Roman Urdu resources, which hinder researchers from creating NLP tools that can reliably handle this language variant for information retrieval, machine translation, multilingual education, and other applications. Current datasets primarily focus on the formal script, with only a few small datasets available for Roman Urdu, leaving a substantial gap in resources for the digital Latin-script form of Urdu that now drives everyday communication among Urdu speakers worldwide.

This study fills this gap by presenting a novel, artificial 75,146-sentence parallel corpus from Roman Urdu to English. We produced this dataset using a hybrid methodology that blends real-world discussions from WhatsApp groups made up of fluent English and Roman Urdu speakers with prompt engineering techniques. We aim to capture a wide range of Roman Urdu-specific linguistic features, contextual subtleties, and colloquial usage by combining synthetic and real-world data. We prioritized diversity during the dataset construction process, ensuring the inclusion of various sentence structures, including code-switching with English and different translations for the same words, thus offering a comprehensive resource for language processing applications.

This corpus holds significant potential to advance low-resource language machine translation and natural language processing research. It supports educational initiatives by enabling researchers to identify and create resources that promote Urdu-English bilingualism and intercultural understanding. By overcoming linguistic barriers, this study advances broader social goals like multilingual education and cross-cultural exchange, as well as specialized fields like natural language processing and machine translation. This paper presents the dataset creation process, corpus analysis, insights, and potential uses of this resource in technical and educational fields.

\section{Related Work}

Roman Urdu, the Latin-script form of Urdu, has become a popular communication medium among younger generations in Urdu-speaking regions, aligning with a global trend of using the Latin script for informal digital communication. While researchers extensively study traditional Urdu, they have paid limited attention to Roman Urdu in natural language processing (NLP). Most researchers focused on sentiment analysis, exploring emotional content in Roman Urdu text\cite{mehmood2018rusad,sharf2017romanurdu,razi2024cyberbullying,zahid2020reviews}
; yet these efforts rely on small datasets, limiting their scope. A Roman Urdu-English parallel corpus could address these gaps, supporting broader NLP applications like sentiment analysis and translation. Despite the increasing prevalence of Roman Urdu, existing parallel corpora predominantly target Roman Urdu-Urdu translation \cite{alam2021romanurduparl,jawaid2011wordorder}, overlooking the broader potential of Roman Urdu-English translation. Researchers recognize that parallel datasets are essential for building machine translation systems that support language learning and cross-cultural communication. For low-resource languages like Roman Urdu, parallel corpora enable translation into high-resource languages like English, where researchers have developed extensive NLP tools and resources. This study addresses the gap by constructing a Roman Urdu-English corpus, providing a critical resource to enhance NLP applications for Roman Urdu text.

Researchers consider parallel data essential for machine translation, enabling systems to learn language translations across languages. The field has evolved significantly, from early statistical machine translation (SMT) methods to neural machine translation (NMT) systems that leverage neural networks for more accurate and flexible translation patterns \cite{ganesh2023survey,pa2016underresourced,yang2020nmt}
, The emergence of transformer-architecture-based, large language models (LLMs) like BERT and T5 has further revolutionized multilingual translation \cite{devlin2018bert,raffel2020exploring}. Addressing this specific challenge of Roman Urdu to English translation, Ahmed et al.(2024)\cite{soomro2024spellingvariations} made a breakthrough contribution by developing two crucial datasets for Roman Urdu-English NLP, a comprehensive collection of 5,244 unique Roman Urdu words with their spelling variations, and a groundbreaking dataset of 35,139 Roman Urdu reviews paired with respective English translations, establishing a strong foundation for future research in this understudied field of language pairs.

In low-resource NLP, synthetic data generation has emerged as a solution to the problem of data scarcity. One method involves choosing sentences that closely resemble in-domain test data from sizable monolingual corpora. This method creates a hypersphere around the centroid for in-domain sentences using continuous vector-space representations (CVR) and selects sentences that fall inside it\cite{chinea2017synthetic}. Artificial translation units (ATUs) are another method pertinent to neural machine translation (NMT) with limited resources. These units tag and substitute high-frequency words in the target language to produce synthetic pairs without needing external resources or pre-trained models \cite{ngo2022synthetic}. Forward Generation (FG) and Backward Generation (BG) techniques have been proposed to generate data with realistic error patterns for automatic post-editing (APE). BG creates synthetic triplets that enhance APE models by introducing translation errors into clean reference texts, whereas FG partially corrects machine-translated texts \cite{lee2021postediting}.

Studies have demonstrated that using large language models (LLMs) to generate synthetic data can improve model performance, especially in low-resource environments. For example, to sample answer candidates for the MRQA task, \cite{schmidt2024fewshot} used Named Entity Recognition (NER) in an existing context. A pre-trained language model, like T5, was then used to generate corresponding questions. The model is then presented with a template that includes the context and the sampled response to generate a question-and-answer pair.  They ensured the quality of the generated questions through rigorous filtering processes, including rule-based filtering and consistency checks, which eliminate irrelevant questions, repeat answers, or remain incomplete. \cite{kaddour2023textaugmentation} also used fine-tuning of a larger LLM, such as GPT-NeoX-20B, on a small labeled dataset to produce more training examples.

Building on these studies, our work utilizes prompt engineering techniques \cite{daimi2024scalabledatadrivenframeworksystematic} to generate synthetic data, creating a Roman Urdu-to-English parallel corpus. This approach not only addresses the data scarcity in Roman Urdu but also supports a broader range of NLP applications, from machine translation to sentiment analysis, enhancing resources for this low-resource language pair.

\section{Dataset Construction}
\begin{figure*}[htbp]
\centering
\vspace{-2cm}
\hspace*{-0.082\textwidth} 
\includegraphics[width=1.2\textwidth, height=0.6\textheight]{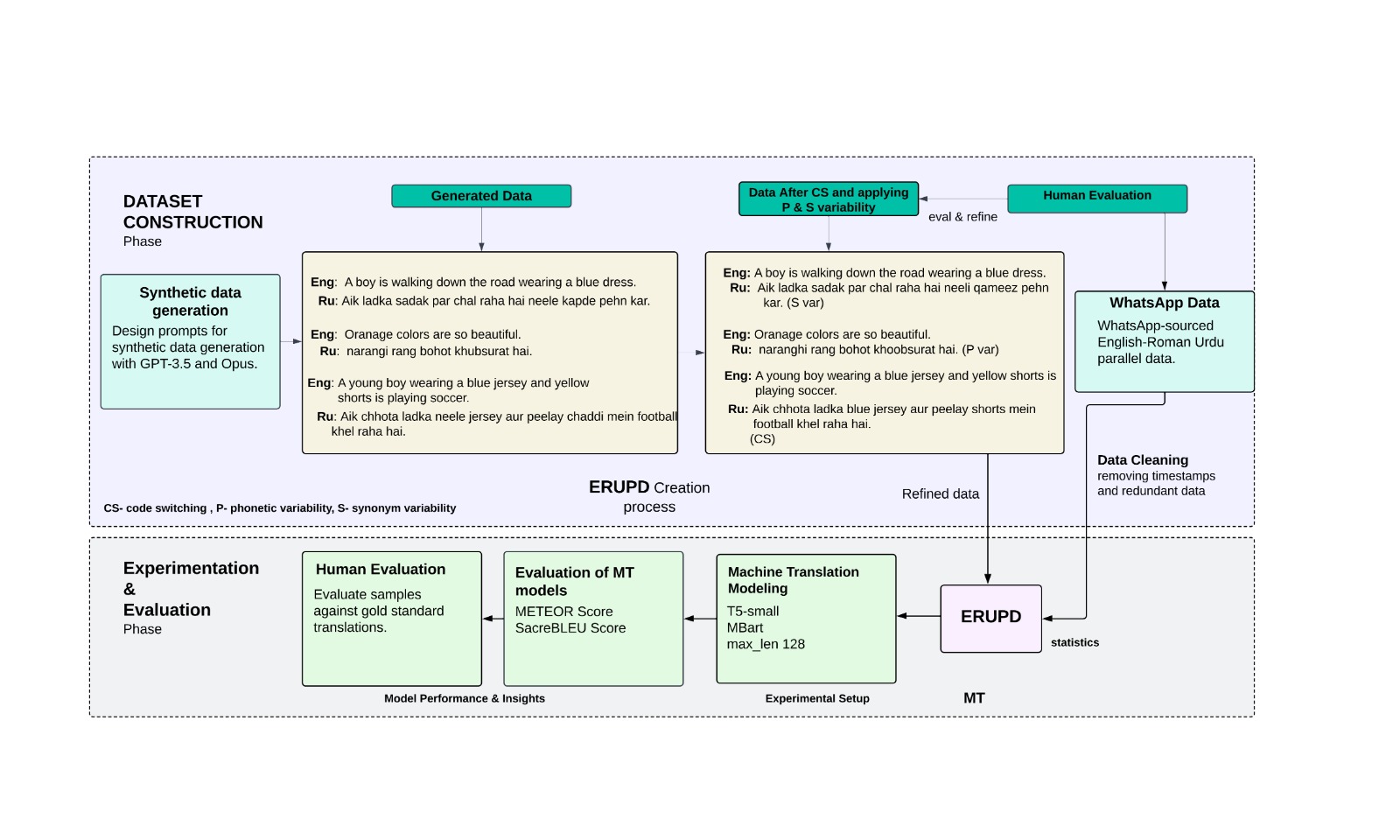}
\vspace{-2.2cm} 
\caption{\large Framework of the process} 
\label{fig}
\end{figure*}

Figure \ref{fig}  illustrates how we constructed the Roman Urdu-to-English parallel corpus. It outlines how we gathered data sources, generated synthetic data through prompt engineering, and applied evaluation and analysis criteria.

\subsection{Data Sources}\label{AA}
We utilized two primary data sources for our dataset: real-world conversations and synthetic data produced through prompt engineering. This hybrid approach enabled us to capture a wide variety of linguistic features, including authentic conversational expressions and controlled syntactic structures. We employed large language models (LLMs), specifically GPT-3.5, GPT-3.5 Turbo Instruct, and Claude Opus, to generate the synthetic data. Additionally, we collected conversational data from a subset of volunteers in WhatsApp groups.

\subsection{Synthetic Data Generation through Prompt Engineering}

We employed several advanced prompt engineering techniques to authentically capture the linguistic diversity and expressive richness of Roman Urdu for the Roman Urdu-to-English parallel corpus. We selected techniques that replicated structural, contextual, and stylistic nuances, ensuring the dataset reflects natural usage across diverse scenarios. Among these techniques, few-shot learning played a critical role. By providing a small set of contextual examples within the prompts, we guided the model to generate sentences exhibiting narrative flow, idiomatic expressions, and diverse vocabulary. These prompts successfully translated English sentences into Roman Urdu and created a wide range of Roman Urdu sentences on diverse topics. 

We also developed topic-specific prompts to generate text tailored to specific themes or linguistic styles, including imitating an author’s style. This approach allowed us to produce sentences that were both contextually rich and stylistically tailored \cite{brown2020fewshot}.

In addition, we used zero-shot learning to generate generalized Roman Urdu sentences without specific contextual examples. This method focused on simplicity and generality, particularly for direct translations. However, we noticed that the model occasionally hallucinated outputs by misinterpreting Roman Urdu as Urdu in its native script, which led to errors. Despite this, zero-shot learning effectively produced simple translations that required minimal contextual information \cite{kojima2022zeroshot}. 

To generate dialogue-based sentences, we applied chain-of-thought prompting, which guided the model to capture sequential reasoning. This technique effectively produced coherent and contextually appropriate outputs requiring emotional nuance and depth. Step-by-step reasoning enabled the model to preserve conversational flow and intricate expressions characteristic of Roman Urdu dialogues\cite{wei2023chainofthought}.

To ensure cultural relevance and authenticity in Roman Urdu usage, we developed a diversified set of prompts covering topics ranging from universal themes to culturally specific scenarios. We selected topics to resonate with Roman Urdu speakers, reflecting everyday interactions, cultural expressions, and contextually rich phrases.

Table \ref{tab1} highlights the two-part composition of these prompts. The main prompt provided a basic template, while the `prompt\_text` section introduced variation to accommodate specific linguistic and contextual demands. For further insight into the main prompt and the variations (`prompt\_texts`) built on them, Appendix \ref{appendix2} offers an in-depth analysis of their design and usage. We organized the outputs generated by the model into two variables, `English\_Sentences` and `Roman\_Urdu\_Sentences`, ensuring clarity and consistency across the dataset.

\begin{table}[htbp]
\caption{Prompt Structure}
\begin{center}
\begin{tabular}{|l|p{5cm}|}
\hline
\textbf{Component} & \textbf{Description} \\
\hline
\textbf{MAIN\_PROMPT} & 
\texttt{'You are an expert linguist in both English and Roman Urdu language (Roman Urdu is Urdu written as English) and you are currently creating a parallel dataset for a neural machine translation model for your paper. Each sentence of the generated content must appear in both languages (English and Roman Urdu), ensuring that they are parallel to each other. The focus is on creating a balanced, rich, and coherent dataset.'} \\
\hline
\textbf{Prompt\_texts (Example)} & 
\texttt{'Generate a children’s story titled 'Thomas the Train.' Include imaginative dialogues and moral lessons, with Roman Urdu translations for each sentence.'} \\
\hline
\textbf{Output Variables} & 
The output is structured into two variables: \textit{English\_sentences} and \textit{Roman\_Urdu\_sentences}, ensuring clarity and consistency across the dataset. \\
\hline
\end{tabular}
\label{tab1}
\end{center}
\end{table}

\subsection{Data Refinement}
After generating the data, we noticed that the dataset exhibited code-switching, phonetic variability, and synonym variability, but these appeared in proportions that were either excessive or insufficient, posing challenges. To address this, we designed prompts with specific instructions to control the desired amount of code-switching, phonetic variability, and synonym usage. However, we observed that the generated sentences often did not fully adhere to these instructions. As a result, we made all adjustments for these linguistic aspects during the human evaluation phase of the dataset.

\subsubsection{Code Switching}
Code-switching between Roman Urdu and English is a common linguistic feature among bilingual speakers. To reflect this, we designed prompts that integrated English phrases within Roman Urdu syntax, capturing the natural blending characteristic of code-switched Roman Urdu. During refinement, we adjusted the inclusion or removal of English elements to ensure linguistic consistency. This process enhanced the authenticity of code-switching in the dataset. Table \ref{tab2} highlights the types of code-mixing we achieved. Examples such as \textit {"Aik chhota ladka neela jersey aur peelay shorts mein football khel raha hai"} demonstrate how English terms such as "jersey," and "football" are seamlessly blended into Roman Urdu sentence structures.

\begin{table}[htbp]
\caption{Code Switching in the dataset}
\begin{center}
\begin{tabular}{|p{4cm}|p{4cm}|}
\hline
\textbf{English} & \textbf{Roman Urdu} \\
\hline
A young boy wearing a blue jersey and yellow shorts is playing soccer. & 
\textit {Aik chhota ladka neela jersey aur peelay shorts mein football khel raha hai.} \\
\hline
A happy woman is preparing a refreshment at a coffee shop. & 
\textit {Aik khush aurat aik coffee ki dukaan mein refreshment tayyar kar rahi hai. }\\
\hline
A person dressed in a blue coat is standing on a busy sidewalk, studying a painting of a street scene. & 
\textit {Aik shakhs jo neela coat pehn kar ek masroof footpath par khada hai, aik sadak ke manzar ki painting ko padhta hua.} \\
\hline
A group of workers on a boardwalk wearing fluorescent vests, holding light wands. & 
\textit {Aik tabqa mazdooron ka boardwalk par fluorescent vests pehne hue light wands pakad kar hai. }\\
\hline
Rafts and a helicopter over water. & 
\textit {Rafts aur ek helicopter paani ke upar hai.} \\
\hline
\end{tabular}
\label{tab2}
\end{center}
\end{table}

\subsubsection{Phonetic Variability}
Phonetic variability is a challenge in Roman Urdu due to inconsistent spelling practices and the lack of a standardized script. Speakers often use different spellings for the same sounds based on personal or regional preferences. To address this, we enriched the dataset during human evaluation phase by adding multiple phonetic representations for words. This allowed us to capture a broad spectrum of variations, reflecting dialectal and personal differences.  For example, we included both  \textit {"Naranghi"} and \textit {"Narangi"} as representations of the word “orange” (see Table \ref{tab3}).

\begin{table}[htbp]
\caption{Phonetic Variability of Roman Urdu Data}
\begin{center}
\begin{tabular}{|p{4cm}|p{4cm}|}
\hline
\textbf{English Form} & \textbf{Roman Urdu Variants} \\
\hline
Orange & \textit {Naranghi, Narangi} \\
\hline
Man & \textit {Admi, Aadmi} \\
\hline
Small & \textit {Chhota, Chota} \\
\hline
Hammer & \textit {Hatora, Hatoda} \\
\hline
Little & \textit {Thoda, Thora} \\
\hline
Ready & \textit {Taiyyar, Tayyar} \\
\hline
\end{tabular}
\label{tab3}
\end{center}
\end{table}

\subsubsection{Synonym Variability}
Roman Urdu, like many spoken and informal language variants, exhibits a rich array of synonymous expressions that reflect the linguistic diversity within Urdu dialects and individual speaker preferences. We incorporated a range of synonymous terms into the dataset to represent this variability accurately. For instance, we included \textit {"Nojawan"} and \textit {"Jawan"} for “young” and \textit {"Kapde"}, \textit {"Qameez"}, and \textit {"Libaas"} for “dress” (see Table \ref{tab4}). This inclusion allowed our dataset to have a more nuanced understanding of Roman Urdu, reflecting the flexibility and richness of everyday communication.

\begin{table}[htbp]
\caption{Synonym Variability in Roman Urdu Data}
\begin{center}
\begin{tabular}{|p{4cm}|p{4cm}|}
\hline
\textbf{English} & \textbf{Roman Urdu Synonyms} \\
\hline
Young & \textit {Nojawan, Chhota, Jawan} \\
\hline
Dress & \textit {Kapde, Qameez, Libaas, Poshak} \\
\hline
Busy & \textit {Masroof, Mashghool, Bhari} \\
\hline
Road & \textit {Sadak, Raasta} \\
\hline
Man & \textit {Aadmi, Mard} \\
\hline
Old man & \textit {Budha, Buzurg, Budhape mei} \\
\hline
\end{tabular}
\label{tab4}
\end{center}
\end{table}

\vspace{-0.5cm}
\subsection{Other Data Sources}
We enhanced the corpus by integrating an existing open-source Roman Urdu sentiment analysis dataset from Kaggle, which comprises approximately 10,000 sentences. We preprocessed this dataset by removing emojis and informal symbols, then translated the Roman Urdu sentences into English and added them to the ERUPD corpus \cite{sharf2017romanurdu}. 

To enhance our dataset, we established two WhatsApp groups with volunteer participants—one group focused on Roman Urdu and the other on English. These volunteers, mostly students who frequently
communicate in both languages, engaged in conversations on
various informal topics. This real-world conversational data
enriched the dataset by reflecting the natural usage patterns,
informal expressions, and conversational dynamics of Roman
Urdu in digital contexts.

\subsection{Challenges and Human Evaluation}\label{SCM}
While developing the English-to-Roman Urdu parallel corpus, we encountered several challenges, particularly with translations produced through prompt engineering. Large language models (LLMs) frequently swapped masculine and feminine forms, leading to inaccurate gender representation. For example, they mixed up Roman Urdu verbs like  \textit {"rehta"} (masculine) and \textit {"rehti"} (feminine) causing contextual errors. We also noticed confusion between singular and plural verb forms, where the model used singular verbs like \textit {"tha"} instead of plural forms like  \textit {"the"} and vice versa. These issues occurred repeatedly because Roman Urdu’s informal and phonetic nature lacks standardized grammatical rules. Additionally, we identified instances where the model hallucinated, failing to provide translations for certain sentences.

We identified and resolved these problems during the human evaluation phase, improving the dataset’s linguistic and contextual accuracy. For instance, we fixed gender misidentifications like \textit {"Usne kaha ke woh har din kaam karta hai"} by changing it to \textit {"Usne kaha ke woh har din kaam karti hai"} to match the feminine subject. Similarly, we corrected plural-singular discrepancies, such as changing \textit {"Woh saare log wahan tha"} to \textit {"Woh saare log wahan the"}. In cases of hallucination, we manually generated the correct translations, ensuring the dataset was complete. We also addressed issues like code-switching, phonetic variability, and synonym variability during this phase, ensuring the dataset effectively captured these linguistic features.

\section{Dataset Statistics}

After creating the dataset, we computed descriptive statistics, revealing that it contains 75,146 parallel sentence pairs in English and Roman Urdu. The average sentence length is 15.81 words in English and 19.03 words in Roman Urdu, reflecting the natural differences in sentence structure and phrasing between the two languages. 
We found that the vocabulary sizes are 60,994 for English and 76,100 for Roman Urdu, indicating a comparable lexical richness in both languages. In terms of translation consistency, the average translation length ratio is 1.20 from Roman Urdu to English and 0.88 from English to Roman Urdu, which captures the natural tendency for Roman Urdu expressions to be longer.
Our total token count is 1,188,049 in English and 1,430,038 in Roman Urdu. This reflects the prevalence of common conversational terms such as   \textit {“ke,” “aur,”} and \textit {“mein”} in Roman Urdu and “a,” “and,” and “the” in English. These statistics demonstrate that the dataset is well-suited for natural language processing applications, particularly those focused on bilingualism and Roman Urdu language processing.

\section {Experiments \& Results}
To demonstrate the effectiveness of the English-to-Roman Urdu parallel corpus, we carried out experiments utilizing two transformer-based neural machine translation (NMT) models: T5-Small and mBART. These experiments evaluated how well the dataset served as a benchmark for machine translation systems in Roman Urdu. 

\subsection{Experimental Setup \& Modeling}
We divided the dataset into training (80\%), validation (18.5\%), and test (1.5\%) sets, ensuring the test set comprised approximately 200 samples. During preprocessing, we addressed inconsistencies in Roman Urdu spellings and transliterations by performing tokenization and normalization. We standardized the maximum sequence length to 128 tokens for tokenization across both models to maintain uniformity.

To tailor the models for Roman Urdu, we customized the mBART model (Liu et al., 2020)\cite{liu2020multilingualdenoisingpretrainingneural} by incorporating a \texttt{<roman\_urdu>} token, optimizing its output for Roman Urdu translations. Additionally, we fine-tuned the T5-Small model (Raffel et al., 2020)\cite{raffel2020exploring} on the dataset to produce high-quality translations. Using the Hugging Face Transformers library, we leveraged the pre-trained capabilities of both models to effectively handle Roman Urdu’s linguistic complexities.

For evaluation, we used SacreBLEU (Post, 2018)~\cite{post2018clarityreportingbleuscores} to measure n-gram precision for lexical overlap. We also applied METEOR (Banerjee and Lavie, 2005)~\cite{lavie2007meteor}, prioritizing recall and semantic equivalence. Additionally, we adapted the METEOR metric to account for Roman Urdu’s unique orthography and phonetics. Table~\ref{tab5} summarizes the SacreBLEU and METEOR scores we achieved, showcasing the models’ translation performance.

\begin{table}[htbp]
\caption{Evaluation Metrics (BLEU and METEOR) for T5-Small and mBART Models}
\begin{center}
\begin{tabular}{|p{2.5cm}|p{2.5cm}|p{2.5cm}|} 
\hline
\textbf{Model (NMT)} & \textbf{BLEU Score (\%)} & \textbf{METEOR Score} \\
\hline
T5-Small & 39.929 & 0.526 \\
\hline
mBART & 38.770 & 0.531 \\
\hline
\end{tabular}
\label{tab5}
\end{center}
\end{table}

\subsection{Model Performance and Insights}
After modeling we reviewed the translations produced by the models and found comparable performance overall, with occasional variations. Notably, the mBART model showed slightly superior performance in specific cases, particularly when handling informal expressions and culturally specific phrases. This qualitative analysis suggested that mBART’s pre-trained multilingual capabilities made it more adept at capturing certain subtleties of Roman Urdu. Furthermore, mBART outputs often resembled gold-standard translations more closely, reflecting its strength in aligning with the linguistic and contextual nuances of the dataset.

To provide deeper insights into how the dataset performs in machine translation tasks, we included examples of outputs generated by the T5-Small and mBART models in Appendix \ref{appendix1} These samples illustrate how each model handles various linguistic aspects of Roman Urdu, including informal spellings, cultural influences, and complex sentence structures. Readers can refer to the appendix to explore the translated texts in detail and evaluate the models' effectiveness on the dataset.

\section{Conclusion and Future Work}
This research introduces a new parallel dataset in both English and Roman Urdu, acting as an essential resource for advancing machine translation tasks and other natural language processing applications related to Roman Urdu. By implementing a thoughtfully constructed pipeline, the dataset ensures its reliability applies to a variety of linguistic contexts, by considering the various features of Roman Urdu, such as code-switching, phonetic variability, and synonym diversity. The effectiveness of the dataset was illustrated by evaluating two models, T5-Small and mBART, which showed that the dataset can effectively facilitate high-quality translation tasks.

Further future work can be done to increase the diversity and representative samples of the dataset, which can enhance its utility for different NLP applications. In addition, this dataset opens ways for investigating linguistic phenomena, like code-switching and informal use of language, in Roman Urdu, hence providing valuable information on what makes the language unique. A more ambitious attempt at creating standardized datasets for low-resource especially those expressed in Latin script or Romanized forms, would significantly benefit the research community by advancing inclusivity and linguistic diversity in NLP research.

\newpage
\onecolumn
\appendix
\section*{Section 1: Illustrative Examples of Model Outputs}\label{appendix1}
\begin{table}[h!]
\centering
\caption{Illustrative Examples of Model Outputs}
\renewcommand{\arraystretch}{1.4} 
\setlength{\tabcolsep}{4pt} 
\begin{tabular}{|p{4.2cm}|p{4.2cm}|p{4.2cm}|p{4.2cm}|}
\hline
\textbf{English Source} & \textbf{Reference (Roman Urdu)} & \textbf{T5-Small Output} & \textbf{mBART Output} \\ \hline
What is your name? & \textit{Apka naam kya hai?} & \textit{Apka naam kya hai?} & \textit{Aap ka naam kya hai?} \\ \hline
Arsenal vs Spurs is a proper Football match & \textit{Arsenal vs Spurs aik zabardast football match hai} & \textit{Arsenal vs Spurs aik sahi football match hai.} & \textit{Arsenal vs Spurs sahi Football match hai.} \\ \hline
It is better for students to follow their dreams & \textit{Talba ke liye behtar hai ke woh apne khwabon ka peecha kare.} & \textit{Talib ilm ko apne khwabon ko follow karna behtar hai.} & \textit{Behtar hai talba ke liye apne khwabon ko follow karen.} \\ \hline
Work hard for something and you will reach success & \textit{Kisi cheez ke liye mehnat karo aur aap ko kamiyabi mile gi.} & \textit{Kuch cheez ke liye mehnat karo aur aap kamiyabi hasil karo.} & \textit{Mehnat kar ke aap kamiyabi tak pohanch jayenge.} \\ \hline
I live in Cheltenham & \textit{Mei Cheltenham mein rehta hoon.} & \textit{Main Cheltenham mein rehta hoon.} & \textit{Main Cheltenham mein rehta hoon.} \\ \hline
The goal of my life is to bring justice to the people and bring about a positive change & \textit{Meri zindagi ka maqsad logon ko insaaf dilana aur ek musbat tabdeeli laane ka hai.} & \textit{Meri zindagi ka maqsad hai ke logon ko insaaf laana aur ek musbat tabdeeli laane ka hai.} & \textit{Meri zindagi ka maqsad logon ko insaaf dena aur ek musbat tabdeeli laane ka hai.} \\ \hline
British people love Indian food & \textit{Angrezi log Hindustani khana pasand karte hain.} & \textit{British log Indian khana pasand karte hain.} & \textit{British log Hindustani khana pasand karte hain.} \\ \hline
\end{tabular}
\label{tab6}
\end{table}

Table~\ref{tab6} presents sample outputs from the T5-Small and mBART models for a subset of test sentences from the English-to-Roman Urdu parallel corpus. Each row includes:
\begin{itemize}
    \item \textbf{Source (English)}: The original English sentence.
    \item \textbf{Reference (Roman Urdu)}: The human-annotated (gold standard) translation.
    \item \textbf{T5-Small Output}: The translation generated by the fine-tuned T5-Small model.
    \item \textbf{mBART Output}: The output from the fine-tuned mBART model.
\end{itemize}

The examples highlight linguistic features such as code-switching, informal phrasing, and cultural nuances in Roman Urdu. 
While both models generally perform well, slight variations in style and word choices reflect differences in their fine-tuning 
and pre-training approaches. This table provides qualitative insights to complement the quantitative evaluations, demonstrating 
the ERUPD dataset’s utility for machine translation tasks.

\clearpage
\appendix
\section*{Section 2: Dataset Generation Prompts}\label{appendix2}
\begin{table}[h!]
\centering
\caption{Main Prompt for Dataset Generation}
\begin{tabular}{|l|}
\hline
\textbf{Main Prompt} \\ \hline
\begin{minipage}{0.94\linewidth}
\vspace{0.2cm}
\texttt{"You are an expert linguist in both English and Roman Urdu language (Roman Urdu is Urdu written as English) and you are currently creating a parallel dataset for a neural machine translation model for your paper. Each sentence of the generated content must appear in both languages (English and Roman Urdu), ensuring that they are parallel to each other. The focus is on creating a balanced, rich, and coherent dataset."}
\vspace{0.2cm}
\end{minipage} \\ \hline
\end{tabular}
\label{tab7}
\end{table}

\renewcommand{\arraystretch}{1.3} 

\begin{table}[h!]
\centering
\caption{Dataset Generation Prompts}
\begin{tabular}{|p{0.2\linewidth}|p{0.3\linewidth}|p{0.4\linewidth}|} 
\hline
\textbf{Objective} & \textbf{Technique} & \textbf{Prompt Text Examples} \\ \hline
\multirow{2}{0.4\linewidth}{\textbf{Story generation}} & \multirow{2}{0.2\linewidth}{Topic-based guidance} & \texttt{\textcolor{blue}{\{MAIN\_PROMPT\}} Generate a novel titled 'The Dragon Warrior,' inspired by Marvel. The story should have more than 100 sentences, 10 paragraphs, and each sentence translated in Roman Urdu.} \\ \cline{3-3}
 &  & \texttt{\textcolor{blue}{\{MAIN\_PROMPT\}} Generate a children’s story titled 'Thomas the Train.' Include imaginative dialogues and moral lessons, with Roman Urdu translations for each sentence.} \\ \hline
\textbf{Conversational dialogues} & Chain of thought & \texttt{\textcolor{blue}{\{MAIN\_PROMPT\}} Write a dialogue between two characters debating a cosmic artifact. Example: English: 'What do you think this artifact is?' Roman Urdu: 'Yeh cheez kya ho sakti hai?' Continue this conversational format.} \\ \hline
\textbf{Scientific topics} & Few-shot learning & \texttt{\textcolor{blue}{\{MAIN\_PROMPT\}} Explain 'The Evolution of Stars' with step-by-step clarity. Example: English: 'Stars form in nebulae, massive clouds of gas.' Roman Urdu: 'Sitaray nebulae mein bante hain jo ke gas ke bade baadal hote hain.' Continue the explanation.} \\ \hline
\multirow{2}{0.4\linewidth}{\textbf{Diverse sentences}} & \multirow{2}{0.2\linewidth}{Few-shot learning} & \texttt{\textcolor{blue}{\{MAIN\_PROMPT\}} Generate diverse sentences on daily activities, include translations. Example: English: 'I love coding on my laptop.' Roman Urdu: 'Main apne laptop par coding karna pasand karta hoon.' Continue with varied scenarios.} \\ \cline{3-3}
 &  & \texttt{\textcolor{blue}{\{MAIN\_PROMPT\}} Generate diverse sentences on food. Include translations. Example: English: 'There is biryani being served in the party.' Roman Urdu: 'Dawat mei biryani parosi jaa rahi hai.' Continue with varied scenarios.} \\ \hline
\textbf{Translate text} & Multi-shot learning & \texttt{\textcolor{blue}{\{MAIN\_PROMPT\}} (complete prompt example) English text to be translated is: Many old females are performing arts and crafts. A crowd of people are in the streets and some have torches. A couple of people have finished shoveling a path. Men working along side a fish and chips restaurant with a dump truck. A man is watching his young daughter's reaction to a homemade birthday cake. People running and walking in and out of a small farm town. A man dressed in black walks on an icy sidewalk. A little girl in a brown top and pink pants is balancing a bowl with plants in it on her head.} \\ \hline
\end{tabular}
\label{tab:dataset_prompts}
\end{table}

\subsection*{Main Prompt Explanation}

The primary prompt showcased in table \ref{tab7} serves as the foundation for developing a bilingual dataset in both English and Roman Urdu. Our main goal is to ensure high-quality translations that maintain precise alignment between the two languages. We pair each English sentence with its Roman Urdu counterpart, which helps us create a balanced and coherent dataset. We use the primary prompt as a prefix to the prompt text and occasionally adjust it to meet the specific requirements of various tasks.

\subsection*{Table Explanation}

The accompanying table \ref{tab:dataset_prompts} provides a detailed overview of the tasks involved in dataset creation. It outlines the objectives, methodologies, and prompt texts associated with each task. We focus each task on specific objectives, ranging from narrative construction to generating everyday sentences. We use methods like topic-based guidance, few-shot learning, chain-of-thought reasoning, and multi-shot learning to enhance the outputs. For example, topic-based guidance focuses on specific themes, while few-shot learning utilizes example-based learning to enhance understanding. Chain-of-thought reasoning fosters clarity by guiding the model through a step-by-step process, and multi-shot learning employs multiple examples to improve content richness and coherence. The prompt texts provide detailed instructions and examples, building on the primary prompt. For instance, storytelling prompts guide the model to create narratives based on widely recognized themes, while scientific prompts ask for thorough explanations.

\subsection*{Content Diversity and Stylistic Variations}

The prompts we created exhibit significant thematic and stylistic diversity, which enriches our ERUPD dataset. The topics span various geographic regions, including the USA, India, and other nations, while also covering abstract themes such as poetry, celestial phenomena, and warfare. This variety adds richness to the dataset. In terms of stylistic variation, we designed narratives that reflect the tonal qualities and nuances of renowned literary figures, such as Enid Blyton and Jane Austen. We created prompts to emulate the distinctive styles of celebrated works, such as those of Sherlock Holmes, Harry Potter, and The Lord of the Rings. Excerpts from these works were integrated into the prompts to generate stylistically similar content.  Additionally, we designed conversational dialogues around intellectually stimulating topics, such as cosmic artifacts, using a stream-of-consciousness style to allow for natural and engaging interactions among characters. We also designed prompts that aimed to generate content inspired by novels like \textit{Secret Seven} and \textit{Famous Five}, as well as works related to Marvel and DC comics, thereby capturing a diverse array of content within the final ERUPD corpus.

\end{document}